\documentclass[letterpaper]{article}
\usepackage[preprint]{aaai2027}
\usepackage[hyphens]{url}
\usepackage{graphicx}
\usepackage{booktabs}
\usepackage{amsmath,amssymb,amsfonts}
\usepackage{natbib}
\usepackage{caption}
\usepackage{multirow}
\usepackage{array}
\usepackage{makecell}

\title{Reflection Steering: Disentangling Reflection from Reasoning in Activation Space for Token-Efficient Inference}

\author{
Hu Jiarui\textsuperscript{\rm 1}\equalcontrib,
Zhiyuan Wen\textsuperscript{\rm 2}\equalcontrib,
Xiaoyun Liu\textsuperscript{\rm 2},
Jiaxing Shen\textsuperscript{\rm 3},
Yu Yang\textsuperscript{\rm 4}
}
\affiliations{
\textsuperscript{\rm 1}School of Computing and Data Science, The University of Hong Kong, Hong Kong SAR, China\\
\textsuperscript{\rm 2}Department of Computing, The Hong Kong Polytechnic University, Hong Kong SAR, China\\
\textsuperscript{\rm 3}School of Data Science, Lingnan University, Hong Kong SAR, China\\
\textsuperscript{\rm 4}Centre for Learning, Teaching and Technology, The Education University of Hong Kong, Hong Kong SAR, China\\
\texttt{tonny\_hujiarui@163.com}; \texttt{zhiyuan.wen@polyu.edu.hk};\\
\texttt{xiaoyun.liu@connect.polyu.hk}; \texttt{jiaxingshen@ln.edu.hk};\\
\texttt{yangyy@eduhk.edu.hk}
}

\newcommand{\best}[1]{\textbf{#1}}
\newcommand{\sd}[1]{{\normalfont\footnotesize\,$\pm$\,#1}}
\newcommand{\rawref}[1]{\textcolor{black!55}{#1}}

\begin{document}
\maketitle

\begin{abstract}
Large reasoning models often produce reasoning traces with verification, revision, and backtracking. When reflection merely re-checks established results, it wastes reasoning tokens and increases latency. Most existing reflection steering methods add a label-derived mean-difference direction across preset layers, but its entanglement with reasoning and length signals destabilizes the accuracy--efficiency trade-off. In this paper, we propose Reflection Steering, a training-free framework for controlling reflection-associated computation within LLMs by disentangling reflection-related activations from general reasoning. Specifically, we contrast reflective and non-reflective hidden states at each LLM layer, denoise the resulting reflection directions with PCA, and orthogonalize them against general-reasoning directions. To limit downstream amplification from early-layer interventions, we calibrate each layer across multiple intervention strengths on a small set, retain only stable layers, and apply bounded projection removal to their residual-stream activations. We conduct extensive experiments across two public benchmarks and three open-weight LLMs against state-of-the-art activation-steering baselines. Results show that Reflection Steering reduces reasoning tokens by 16.9\% on average across six matched settings. Besides, our method further introduces a bounded reflection intervention-strength parameter $\alpha$, enabling deployment-time adjustment to balance token savings, accuracy, and generation stability.
\end{abstract}

\section{Introduction}
Large reasoning models improve reliability by verifying intermediate results, revising earlier steps, backtracking, and changing strategies within extended reasoning traces \citep{wei2022chain,snell2024scaling,deepseekr1}. Yet reflection can outlast its utility. After a conclusion is already supported by the preceding trace, a model may repeat the same checks without changing its answer, increasing token usage, latency, and serving cost and sometimes contributing to overthinking \citep{chen2024overthinking,sui2025survey}, as shown in Figure \ref{fig:toy}. 
We therefore ask whether a large reasoning model can be equipped with a tunable interface for regulating reflection and reducing its associated computation.

\begin{center}
\includegraphics[width=0.99\columnwidth]{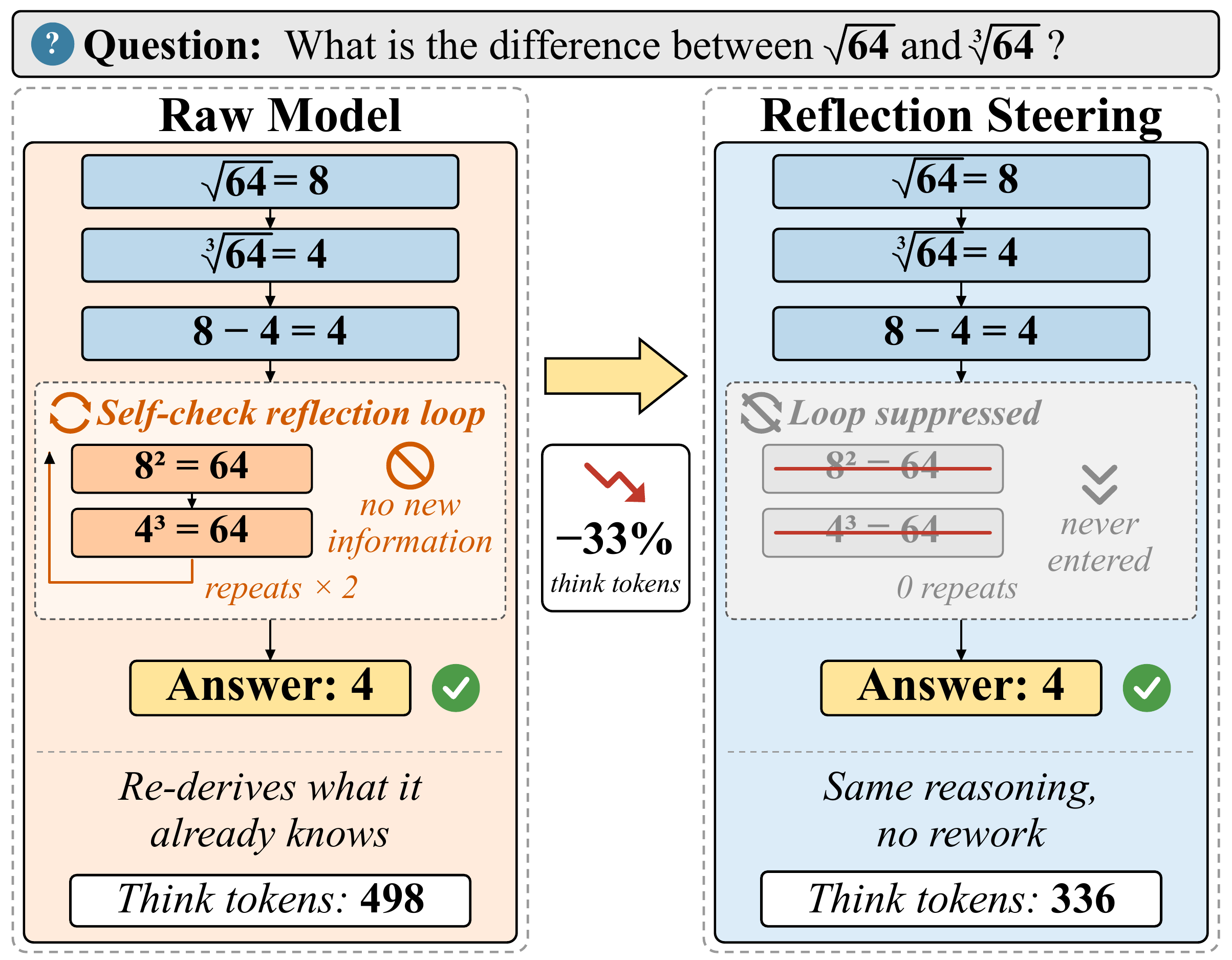}
\captionof{figure}{\textbf{A motivating example of Reflection Steering.} The raw model reaches the correct result but continues to verify it several times. By controlling reflection-associated activations, Reflection Steering produces a shorter trace that reaches the same answer with fewer thinking tokens in this example.}
\label{fig:toy}
\end{center}

Existing approaches reduce reasoning cost through either output-level constraints or representation-level interventions. Output-level methods use token budgets, token pruning, or length-aware training to shorten reasoning traces, but they regulate overall length without distinguishing the computation being removed \citep{muennighoff2025s1,xia2025tokenskip,xu2025chainofdraft,luo2025o1pruner}. Representation-level methods exploit approximately linear residual-stream directions \citep{zou2023repe,turner2023actadd,arditi2024refusal,park2024linear}, with recent extensions targeting overthinking, reflection, and reasoning efficiency \citep{manifold2025,seal2025,reflctrl2025,crest2025,asc2026}. These methods enable targeted control, but depend on a steering direction that isolates the intended behavior. A contrastive estimate may mix reflection-associated signals with reasoning-related structure and other variation, making it difficult to reduce reflection without affecting useful reasoning.

We hypothesize that naive reflection steering may harm useful reasoning because reflection-associated and reasoning-related signals are entangled in activation space. Reflective and non-reflective states differ not only in reflection but also in reasoning content, token position, and trace length. Consequently, their mean difference may combine reflection-associated signals with activation structure shared across reasoning states. Suppressing this mixed direction may therefore reduce reflection while also weakening useful reasoning. This analysis motivates our core intuition: separate reflection-associated signals from shared reasoning-related structure before intervention and focus control on the remaining component.

Building on this intuition, we propose \emph{Reflection Steering}, a training-free representation-level method for reducing reflection-associated computation while limiting interference with general reasoning. Our method addresses three questions: \textbf{what to intervene}, \textbf{where to intervene}, and \textbf{how to intervene}. For \textbf{what to intervene}, we compare layer-wise activations from a small set of reflective and non-reflective examples to estimate an initial reflection direction. We then use PCA to reduce noise and orthogonalization to remove the part aligned with an activation pattern shared by both groups. For \textbf{where to intervene}, we test candidate layers at several strengths on a small calibration set and retain those that consistently reduce reflection-related behavior without destabilizing generation. For \textbf{how to intervene}, we scale down only the component of each token state that lies along the learned direction, rather than applying the same fixed shift to every token. Once calibrated, the direction and layer set remain fixed, while a reflection intervention-strength parameter $\alpha$ controls the degree of projection removal. Model weights, routing parameters, and other decoding settings remain unchanged.

We evaluate Reflection Steering on MATH-500 for mathematical problem solving \citep{hendrycks2021math} and GPQA-Diamond for graduate-level scientific reasoning \citep{rein2024gpqa}. Experiments span three Qwen-family open-weight checkpoints under matched answer parsing and token accounting. We compare against two closely related representation-level baselines: CREST for cognitive-behavior steering \citep{crest2025} and ReflCtrl for reflection control \citep{reflctrl2025}, using their reproduced default operating points. Across six matched model--task settings, Reflection Steering reduces reasoning tokens by 16.9\% on average. On Qwen3-30B-A3B with MATH-500, it reduces tokens by 21.8\% with a $-0.1$-point accuracy change; paired equivalence testing supports equivalence to the raw model within a $\pm1$-point margin. Overall, the strongest accuracy-preserving evidence appears on mathematical reasoning. To summarize, our contributions are:

\begin{itemize}
\item We show that raw mean-difference reflection directions retain substantial overlap with a rank-one proxy for activation structure shared across reasoning states, and that proxy-based orthogonalization reduces this measured overlap to the isotropic-random level.

\item We propose \emph{Reflection Steering}, a training-free activation-space method combining low-rank filtering, proxy-based orthogonalization, layer calibration, and state-dependent projection attenuation. A deployment-time parameter $\alpha$ adjusts intervention strength without updating model weights or decoding settings.

\item Across six model--benchmark pairs spanning three LLMs in different sizes and two benchmarks on math reasoning and science MCQ, Reflection Steering reduces thinking tokens by 16.9\% on average while largely preserving task performance.
\end{itemize}

\section{Related Work}
\paragraph{Controlling reasoning cost outside the residual stream.}
One cluster reduces cost via the prompt, decoding, or weights. Prompt-budget and length-instruction methods ask the model to reason within a token budget \citep{muennighoff2025s1,han2024tale,xu2025chainofdraft}, but reasoning models often ignore such instructions as the chain of thought grows \citep{fu2025scaling}. Early-exit, token-skipping, and stopping rules cut cost directly \citep{xia2025tokenskip,adaptthink2025} at the risk of removing reasoning that is still needed, and training- or RL-based approaches reshape the reasoning policy itself \citep{luo2025o1pruner} but require weight changes and cannot attach to a frozen checkpoint. CGRS suppresses reflection triggers using model confidence \citep{cgrs2025}, operating at the token level rather than on the underlying representation. Our method is complementary: training-free, leaving prompt and decoding untouched, and acting on the representation that produces reflection.

\paragraph{Activation steering for reflection and efficiency.}
Another cluster edits internal activations. Representation-engineering studies show behaviors often align with approximately linear directions modifiable at inference time \citep{zou2023repe,turner2023actadd,li2023iti,rimsky2024caa,arditi2024refusal,park2024linear}, and projection-based concept erasure gives a related geometric view via nullspace projections \citep{ravfogel2020null,belrose2023leace}. Manifold Steering projects an overthinking direction onto a low-dimensional activation manifold, while SEAL distinguishes execution, reflection, and transition states before intervention \citep{manifold2025,seal2025}. ASC learns a compression direction from paired verbose and concise traces under a distributional constraint \citep{asc2026}, and CREST identifies cognitive attention heads and intervenes on their hidden representations \citep{crest2025}. ReflCtrl is closest to ours \citep{reflctrl2025}; the two differ in the activation granularity used to estimate the direction, the treatment of confounding reasoning signals, the selection of intervention layers, and the form of activation modification. Reflection Steering is designed to improve direction specificity and intervention stability.

\begin{figure*}[t]
\centering
\includegraphics[width=0.99\textwidth]{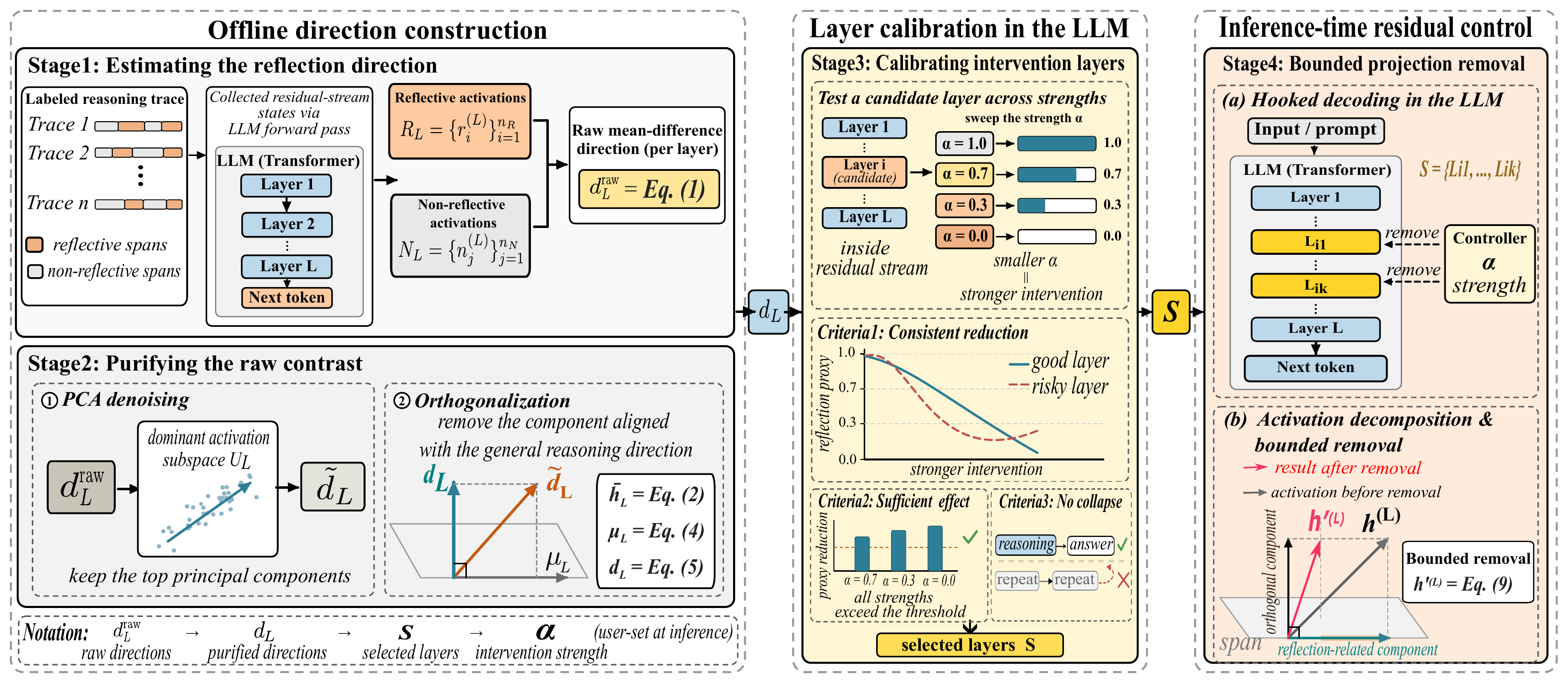}
\caption{\textbf{Reflection Steering overview.} Stage 1 estimates a raw reflection direction from reflective and non-reflective residual states. Stage 2 purifies the direction with PCA denoising and orthogonalization against a general-reasoning direction. Stage 3 tests candidate layers over several intervention strengths and keeps only stable responders. Stage 4 applies bounded projection removal at the selected layers during decoding.}
\label{fig:pipeline}
\end{figure*}

\section{Method}
Reflection Steering turns a noisy reflection contrast into a fixed inference-time controller. Figure~\ref{fig:pipeline} shows the four stages. First, we estimate a raw reflection direction at each layer. Next, we purify the direction so that it is less mixed with general reasoning and sample noise. We then calibrate the candidate layers, because the same intervention can behave differently at different depths. Finally, we apply bounded projection removal only at the selected layers. The model weights and decoding settings remain unchanged. 

\paragraph{Stage 1: Estimating the reflection direction.}
Prior activation-steering methods often represent a behavior with the mean activation difference between positive and negative examples \citep{zou2023repe,turner2023actadd,rimsky2024caa,arditi2024refusal,reflctrl2025}. This gives a simple linear axis in the same residual-stream space where steering is later applied. We use the same idea as the initial estimate of reflection.

For layer $L$, let $R_L=\{r_i^{(L)}\}_{i=1}^{n_R}$ and $N_L=\{n_j^{(L)}\}_{j=1}^{n_N}$ denote activations at reflective and non-reflective positions. The raw direction is
\begin{equation}
 d_L^{\mathrm{raw}}
 =\frac{1}{n_R}\sum_{i=1}^{n_R}r_i^{(L)}
 -\frac{1}{n_N}\sum_{j=1}^{n_N}n_j^{(L)}.
\label{eq:meandiff}
\end{equation}
We use all labeled positions rather than only the first token of each reasoning step. A reflective step usually spans several tokens. Therefore, using the full span makes the estimate less dependent on one boundary token and better represents the activation pattern of the whole step. Still, the mean difference is only a first-order estimate. It can contain signals other than reflection, which motivates Stage 2.

\paragraph{Stage 2: Purifying the raw contrast.}
The two activation groups differ in more than reflective behavior. Their mean difference can also contain prompt and length signals, general reasoning activity, and finite-sample noise. Direct steering along this mixed direction can therefore remove useful information.

We first reduce noise by restricting the raw direction to the main activation subspace. Let
\begin{equation}
 \bar h_L=\frac{\sum_i r_i^{(L)}+\sum_j n_j^{(L)}}{n_R+n_N},
 \qquad
 X_L=\bigl[h-\bar h_L\bigr]_{h\in R_L\cup N_L},
\label{eq:centered}
\end{equation}
where $\bar h_L$ is the mean of all labeled activations at layer $L$. Let $Q_L\in\mathbb{R}^{d\times k}$ contain the leading $k$ right singular vectors of $X_L$. We project the raw direction onto this subspace:
\begin{equation}
 \tilde d_L=Q_LQ_L^{\top}d_L^{\mathrm{raw}}.
\label{eq:pca}
\end{equation}
By the Eckart--Young theorem, this is the optimal rank-$k$ subspace for reconstructing $X_L$ in the least-squares sense \citep{eckart1936lower}. We use it as a low-rank prior on the contrast. As a result, components outside the dominant activation subspace are reduced, while the main structure of the contrast is kept.

PCA alone does not separate reflection from reasoning activity shared by both groups. We therefore use the pooled mean to define a layer-specific general-reasoning direction:
\begin{equation}
 \mu_L=\frac{\bar h_L}{\lVert\bar h_L\rVert_2}.
\label{eq:mu}
\end{equation}
Both reflective and non-reflective positions come from reasoning traces. Thus, their pooled mean gives a simple estimate of the activation component shared by the two groups. It is only a rank-one proxy, not a complete model of general reasoning.

Next, we remove the part of $\tilde d_L$ aligned with this shared direction:
\begin{equation}
 d_L=
 \frac{(I-\mu_L\mu_L^{\top})\tilde d_L}
 {\left\lVert(I-\mu_L\mu_L^{\top})\tilde d_L\right\rVert_2}.
\label{eq:ortho}
\end{equation}
Before normalization, this is the orthogonal projection of $\tilde d_L$ onto the nullspace of $\mu_L^{\top}$. In other words, it is the closest vector to $\tilde d_L$ with zero linear overlap with the estimated shared direction. This operation follows the geometry of projection-based concept erasure \citep{ravfogel2020null,belrose2023leace}. Our claim is narrower: it removes one measured source of entanglement rather than fully separating reflection from all general reasoning.

\begin{figure}[t]
\centering
\includegraphics[width=0.9\columnwidth]{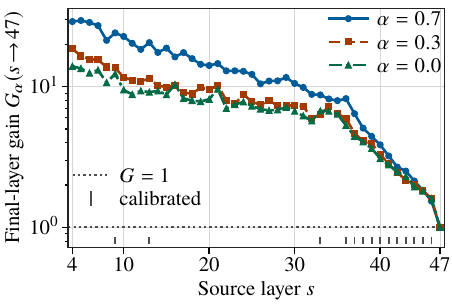}
\caption{\textbf{Downstream amplification of residual-stream interventions.} Median final-layer gain over five teacher-forced MATH traces; ticks below $G=1$ mark calibrated layers.}
\label{fig:propagation}
\end{figure}

\paragraph{Stage 3: Calibrating intervention layers.}
Even a small residual-stream edit can change later computation. Let $F_{s\to k}$ denote the computation from source layer $s$ to a later layer $k$. For a small intervention $\delta h_s$,
\begin{equation}
 \delta h_k
 =F_{s\to k}(h_s+\delta h_s)-F_{s\to k}(h_s)
 \approx J_{s\to k}\,\delta h_s,
\label{eq:propagate}
\end{equation}
where $J_{s\to k}$ is the downstream Jacobian along the current trajectory. Since its norm need not be below one, a local edit can persist or grow through later layers, and edits at several layers can accumulate. We verify this with a teacher-forced audit on five held-out MATH traces. We hold each 2,048-token prefix fixed and intervene at one source layer at a time. For source layer $s$ and downstream layer $k$, we measure
\begin{equation}
 G_{\alpha}(s\!\to\!k)
 =\frac{\lVert\Delta h_k(s,\alpha)\rVert_2}
 {\lVert\Delta h_s(s,\alpha)\rVert_2+\varepsilon},
\label{eq:gain}
\end{equation}
where $\Delta h_k$ is the difference between the intervened and unmodified residual states. For every tested non-baseline $\alpha$, 43 of 44 source layers have a final-layer gain above one, and earlier interventions tend to be amplified more strongly (Figure~\ref{fig:propagation}). Thus, both intervention depth and strength matter. We therefore calibrate layers instead of steering all layers in the same way.

We calibrate each candidate layer on a small calibration subset. Let $m_0(q)$ be the raw-model reflection-proxy count for example $q$, and let $m_{L,\alpha}(q)$ be the count when only layer $L$ is intervened. The mean reduction is
\begin{equation}
 r_L(\alpha)=\frac{1}{|\mathcal C|}\sum_{q\in\mathcal C}
 \bigl[m_0(q)-m_{L,\alpha}(q)\bigr].
\label{eq:calib}
\end{equation}
A layer is retained only when three conditions hold. First, the reduction should remain approximately monotonic as $\alpha$ decreases, so a stronger intervention gives a consistent response. Second, every tested non-baseline strength must produce at least a small positive reduction. This avoids selecting layers whose measured effect is only proxy noise or is locally negligible, because even an unhelpful local edit can still propagate downstream. Third, the calibration runs must not trigger generation collapse. These checks address three different risks: an inconsistent response, an ineffective local edit, and unstable generation. 

The selected layers are therefore reliable only within the tested range. Calibration tests one layer at a time and uses a reflection proxy. It does not certify final-answer accuracy, and it does not guarantee that the strongest intervention across several selected layers will preserve general reasoning. We evaluate those properties on the complete multi-layer controller.

\paragraph{Stage 4: Bounded projection removal.}
The final choice is how to change the activation. Additive controllers such as ReflCtrl apply a fixed displacement along a steering direction \citep{reflctrl2025}. The same update is added even when the current activation contains little reflection-related signal. A large fixed shift can therefore move the state away from its current representation path. Our goal is suppression rather than injection, so we remove only the component that is already present along the purified direction.

For a selected layer $L$ with unit direction $d_L$ and current activation $h^{(L)}\in\mathbb{R}^d$, we apply
\begin{equation}
 h'^{(L)}=h^{(L)}-(1-\alpha)d_Ld_L^{\top}h^{(L)},
\label{eq:steer}
\end{equation}
where $\alpha\in[0,1]$ parameterizes the \emph{reflection intervention strength}. The setting $\alpha=1$ leaves the activation unchanged, whereas smaller values apply a stronger intervention by retaining less of the current projection onto $d_L$. Thus, the update adapts to the current state: it is small when the reflection-related component is small and larger when that component is strong.

For unit $d_L$, the operator has eigenvalue $\alpha$ along $d_L$ and eigenvalue $1$ on its orthogonal complement. Therefore,
\begin{equation}
 \lVert h'^{(L)}\rVert_2\leq\lVert h^{(L)}\rVert_2.
\label{eq:norm}
\end{equation}
This bound applies to the local edit. It does not guarantee that the downstream effect or the final sequence length will change monotonically, because the perturbation can still propagate through later layers as shown in Equation~\ref{eq:propagate}.

The operator is applied at every decoding step only to the selected layers. Model weights, routing parameters, sampling settings, and answer parsing remain unchanged. The projection form builds on prior work on linear representation directions. Our contribution lies in how the direction is estimated, purified, calibrated by layer, and then used for bounded removal.

\section{Experimental Setup}
To validate the method, we organize the setup around the experimental design and the implementation details.

\paragraph{Benchmarks.}
We evaluate on MATH-500 \citep{hendrycks2021math} and GPQA-Diamond \citep{rein2024gpqa}. Exact normalized-record auditing shows that the 150 direction-learning inputs are MATH-500 records 0--149, with the 20 calibration inputs nested inside them, and zero exact overlap with GPQA-Diamond. Full MATH-500 is therefore in-sample for direction construction, so we additionally report a record-disjoint \emph{MATH-350} split (records 150--499) that no fitting or calibration record touches; GPQA-Diamond is fit-disjoint by construction. This lets us separate in-sample behavior, record-disjoint sensitivity, and cross-task transfer.

\paragraph{Models and comparison methods.}
The raw model is Qwen3-30B-A3B; Qwen3-8B and QwQ-32B are cross-checkpoint transfers. We focus on Qwen-family open-weight checkpoints because they provide the internal activation access required by our method while supporting controlled comparisons across model scales; transfer beyond this family remains untested. As external efficiency comparators under identical accounting, we reproduce two representative activation-level controllers: CREST on Qwen3-30B-A3B and ReflCtrl on all three models, each at its default configuration. Comparisons are default-point rather than equalized-tuning frontiers.

\paragraph{Evaluation protocol and metrics.}
We evaluate three aspects. For efficiency we report mean thinking tokens and token reduction against the matched raw model ($\alpha=1$). For task performance we report strict accuracy and, for equivalence testing, a cluster-level paired TOST at a $\pm1$-point margin. For generation stability we report the collapse (repetition-loop) rate and the unparsable (no-box) rate. We summarize the accuracy--cost trade-off with an efficiency ratio $\rho=\Delta\text{Tok}/|\Delta\text{Acc}|$ (higher is better). We use ``---'' for all unavailable or undefined entries; token increases yield negative $\rho$. Reflection markers (verification, uncertainty, backtracking, self-correction, strategy switching) are used only as calibration and interpretability proxies, not as success metrics.

\paragraph{Implementation details.}
All runs use temperature 0.6, top-$p$ 0.95, and a 32{,}768-token cap. The 30B contrast uses reflection-encouraging versus commit-without-revisiting prompts; 8B/32B instead use regex-labeled reflective versus clean steps because their extended-thinking phases are less prompt-responsive \citep{qwq32b,fu2025scaling}. Calibration tests $\alpha\in\{0.7,0.3,0.0\}$ on Qwen3 and a checkpoint-specific grid on QwQ-32B, retaining only layers with monotone proxy reduction, a minimum effect, and no collapse. This selects 14/44, 2/34, and 6 layers on 30B, 8B, and 32B, respectively. For the main Qwen3-30B-A3B comparisons, we keep one direction and layer set fixed and average decoding seeds 0, 64, and 128. Accuracy intervals and TOST use question-cluster bootstraps preserving within-seed pairing; table deltas use unrounded means, whereas $\rho$ uses displayed values. Because $\alpha=0.7$ was selected retrospectively from task summaries, prospective evidence comes from a separate hash-frozen pilot.

\section{Results}
We organize the evaluation around three research questions: whether Reflection Steering improves the accuracy--cost trade-off (RQ1), why each component of the method is necessary (RQ2), and whether the method is stable across tasks, models, and sampling (RQ3).

\subsection{RQ1: Does Reflection Steering improve the accuracy--cost trade-off?}
Table~\ref{tab:main} compares one fixed controller with prior methods under identical parsing and token accounting on MATH-500, record-disjoint MATH-350, and GPQA-Diamond.

\begin{table*}[t]
\centering
\small
\setlength{\tabcolsep}{5.5pt}
\begin{tabular}{llrrrrrr}
\toprule
Benchmark & Method & Tok.\,$\downarrow$ & $\Delta$Tok.\,(\%)\,$\uparrow$ & Acc.\,(\%)\,$\uparrow$ & $\Delta$Acc.\,(pp)\,$\uparrow$ & $\rho$\,$\uparrow$ & Coll.\,(\%)\,$\downarrow$\\
\midrule
\multirow{4}{*}{MATH-500}
 & \rawref{Raw model} & \rawref{4679\sd{67}} & \rawref{---} & \rawref{95.1\sd{0.1}} & \rawref{---} & \rawref{---} & \rawref{0.0}\\
 & CREST & 5052\sd{127} & $-8.0$ & 94.8\sd{0.3} & $-0.3$ & $-26.7$ & 0.2\\
 & ReflCtrl ($\lambda{=}{-}0.48$) & 4031\sd{23} & 13.9 & 94.1\sd{0.8} & $-1.0$ & 13.9 & 0.3\\
 & Reflection Steering & \best{3657}\sd{11} & \best{21.8} & \best{94.9}\sd{0.8} & \best{$-0.1$} & \best{218.0} & \best{0.0}\\
\midrule
\multirow{4}{*}{\shortstack[l]{MATH-350\\(disjoint)}}
 & \rawref{Raw model} & \rawref{4796\sd{75}} & \rawref{---} & \rawref{95.3\sd{0.2}} & \rawref{---} & \rawref{---} & \rawref{0.0}\\
 & CREST & 5192\sd{148} & $-8.3$ & 95.3\sd{0.3} & $0.0$ & --- & 0.0\\
 & ReflCtrl ($\lambda{=}{-}0.48$) & 4109\sd{66} & 14.3 & 94.5\sd{0.8} & $-0.9$ & 15.9 & 0.3\\
 & Reflection Steering & \best{3674}\sd{55} & \best{23.4} & \best{95.4}\sd{0.8} & \best{$+0.1$} & \best{234.0} & \best{0.0}\\
\midrule
\multirow{4}{*}{GPQA-D}
 & \rawref{Raw model} & \rawref{7091\sd{242}} & \rawref{---} & \rawref{71.9\sd{1.1}} & \rawref{---} & \rawref{---} & \rawref{0.0}\\
 & CREST & 6849\sd{52} & 3.4 & 68.5\sd{1.5} & $-3.4$ & 1.0 & 0.2\\
 & ReflCtrl ($\lambda{=}{-}0.48$) & 5836\sd{11} & 17.7 & 70.4\sd{2.0} & $-1.5$ & 11.8 & 0.0\\
 & Reflection Steering & \best{5605}\sd{136} & \best{21.0} & \best{70.4}\sd{0.6} & \best{$-1.5$} & \best{14.0} & \best{0.0}\\
\bottomrule
\end{tabular}
\caption{\textbf{Performance comparison on Qwen3-30B-A3B.} Thinking-token and accuracy values are three-seed means $\pm$ standard deviations. Reflection Steering uses a reflection intervention strength of $\alpha=0.7$; CREST \citep{crest2025} and ReflCtrl \citep{reflctrl2025} use reproduced defaults. Gray raw-model rows are references; bold marks the best controlled value per column, with only Reflection Steering bold on ties. ``---'' denotes an unavailable or undefined entry.}
\label{tab:main}
\end{table*}

Reflection Steering gives the strongest or near-strongest trade-off on all three splits. It cuts thinking tokens by 21.8\% on MATH-500 with a $-0.1$-point accuracy change, and by 23.4\% on record-disjoint MATH-350 with a $+0.1$-point change; CREST instead increases thinking tokens on both MATH splits at its reproduced default. The slight MATH-500 decrease may reflect an accuracy--efficiency cost of reducing thinking tokens, although it remains within the reported $\pm1$-point equivalence margin. On GPQA-Diamond, the same controller transfers without refitting and matches ReflCtrl's mean accuracy while reducing more tokens (21.0\% versus 17.7\%).

\paragraph{Adjustable intervention strength.}
Table~\ref{tab:strength} varies the reflection intervention-strength parameter $\alpha$. Smaller $\alpha$ generally removes more thinking tokens, but the strongest intervention does not always improve compression or stability. We use $\alpha=0.7$ because it captures most of the token saving with zero or near-zero collapse and the best overall trade-off across the three splits. Thus, $\alpha$ provides deployment-time control over the accuracy--cost trade-off rather than a ``stronger is better'' setting.

\paragraph{Trace-dependent compressibility.}
On paired MATH-500 seed-128 traces, raw-model trace length is negatively correlated with proportional token reduction, with Spearman $\rho=-0.264,-0.370,-0.418$ for $\alpha=0.7,0.3,0.0$, respectively. Thus, longer raw-model traces tend to undergo smaller proportional compression, indicating that trace length alone does not determine how much reasoning can be removed by a fixed intervention strength.

\begin{table}[t]
\centering
\small
\setlength{\tabcolsep}{4pt}
\begin{tabular*}{\columnwidth}{@{\extracolsep{\fill}}llrrrr@{}}
\toprule
Benchmark & $\alpha$ & Tok.\,$\downarrow$ & $\Delta$Tok.\,(\%)\,$\uparrow$ & Acc.\,(\%)\,$\uparrow$ & Coll.\,(\%)\,$\downarrow$\\
\midrule
\multirow{4}{*}{MATH-500}
 & \rawref{1.0} & \rawref{4679} & \rawref{---} & \rawref{95.1} & \rawref{0.0}\\
 & 0.7 & 3657 & 21.8 & 94.9 & 0.0\\
 & 0.3 & 3292 & 29.6 & 94.6 & 0.2\\
 & 0.0 & 3439 & 26.5 & 93.5 & 1.1\\
\midrule
\multirow{4}{*}{\shortstack[l]{MATH-350\\(disjoint)}}
 & \rawref{1.0} & \rawref{4796} & \rawref{---} & \rawref{95.3} & \rawref{0.0}\\
 & 0.7 & 3674 & 23.4 & 95.4 & 0.0\\
 & 0.3 & 3216 & 32.9 & 95.5 & 0.0\\
 & 0.0 & 3389 & 29.3 & 94.4 & 0.8\\
\midrule
\multirow{4}{*}{GPQA-D}
 & \rawref{1.0} & \rawref{7091} & \rawref{---} & \rawref{71.9} & \rawref{0.0}\\
 & 0.7 & 5605 & 21.0 & 70.4 & 0.0\\
 & 0.3 & 5437 & 23.3 & 69.7 & 0.2\\
 & 0.0 & 5368 & 24.3 & 68.9 & 0.0\\
\bottomrule
\end{tabular*}
\caption{\textbf{Performance across intervention strengths.} Gray $\alpha=1.0$ rows are raw-model references; the remaining rows report intervened settings within each benchmark under matched accounting. The deployed setting is $\alpha=0.7$.}
\label{tab:strength}
\end{table}

\paragraph{Multi-seed consistency and statistical validity.}
The token savings are consistent across seeds. At $\alpha=0.7$, reductions are 22.5\%, 22.3\%, and 20.7\% on MATH-500 and 22.2\%, 20.9\%, and 19.8\% on GPQA-Diamond, with zero collapse. Paired TOST passes on MATH-500 ($-0.13$ points, 90\% CI [$-0.60$, $+0.33$], $p=0.0012$) and MATH-350 ($+0.10$, [$-0.51$, $+0.70$], $p=0.0074$), supporting equivalence within $\pm1$ point. GPQA-Diamond does not pass the same equivalence test ($-1.52$ points, [$-4.38$, $1.18$]).

\paragraph{Prospective pilot on fresh data.}
We also fix the protocol before generation and validate on 500 fresh, overlap-filtered MATH-train problems. Thinking tokens fall by 21.0\%, with a one-sided 95\% lower bound of 17.6\%, above the pre-registered 10\% target, and neither arm collapses. Accuracy changes by $-0.60$ points, with a one-sided lower bound of $-1.69$, which does not meet the one-point criterion. Overall, RQ1 shows a consistent efficiency gain and accuracy equivalence on the retrospective mathematical evaluations.

\subsection{RQ2: Are purification, layer calibration, and projection removal necessary?}
We next ask which design choices make the method work, by removing each in turn. Because direction construction and layer calibration use MATH-500 records, we conduct these component ablations on the fit-disjoint GPQA-Diamond split to test whether the components remain necessary beyond the construction data. Table~\ref{tab:ablation} reports the ablations at $\alpha=0.7$.

\begin{table}[t]
\centering
\small
\setlength{\tabcolsep}{4pt}
\begin{tabular*}{\columnwidth}{@{\extracolsep{\fill}}lrrrr@{}}
\toprule
Variant & Tok.\,$\downarrow$ & $\Delta$Tok.\,$\uparrow$ & Acc.\,$\uparrow$ & Coll.\,$\downarrow$\\
\midrule
\multicolumn{5}{l}{\emph{Panel A: direction purification}}\\
Raw model & 7250 & --- & 70.7 & 0.0\\
PCA + orthogonalization & 5737 & 20.9 & 70.7 & 0.0\\
No orthogonalization & 4104 & 43.4 & 67.2 & 0.0\\
No PCA & 6105 & 15.8 & 71.2 & 0.0\\
\midrule
\multicolumn{5}{l}{\emph{Panel B: layer calibration and projection removal}}\\
Calibrated + projection & 5737 & 20.9 & 70.7 & 0.0\\
All layers + projection & 5651 & 22.1 & 66.7 & 0.5\\
Calibrated + additive & 9268 & $-27.8$ & 69.2 & 0.0\\
\bottomrule
\end{tabular*}
\caption{\textbf{Ablations on GPQA-Diamond.} Full 198-question split, $\alpha=0.7$.}
\label{tab:ablation}
\end{table}

\paragraph{Purification.}
Table~\ref{tab:ablation}, Panel A tests direction purification. The full method matches raw-model accuracy while cutting tokens by 20.9\%. Without PCA, token reduction falls to 15.8\%. Without orthogonalization, compression increases but accuracy drops to 67.2\%. Thus, PCA improves the direction estimate, while orthogonalization is important for preserving useful reasoning.

\begin{figure}[t]
\centering
\includegraphics[width=0.98\columnwidth]{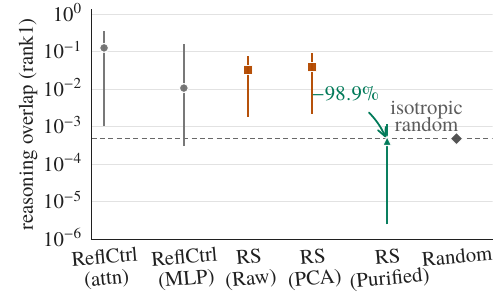}
\caption{\textbf{Reasoning-subspace overlap of steering directions.} Medians over 14 layers; bars span the 5th--95th percentiles. RS is Reflection Steering; RS (Raw) is its unpurified mean-difference direction. Random is an isotropic negligible-overlap reference. PCA retains substantial reasoning-proxy overlap, whereas purification cuts it by 98.8\% to the isotropic-random level. CREST has no comparable single direction.}
\label{fig:entangle}
\end{figure}

\paragraph{Direction specificity.}
Held-out GPQA-Diamond traces confirm this geometry (Figure~\ref{fig:entangle}). We report median rank-one overlap across the 14 deployed layers, with bars spanning the 5th--95th percentile. CREST \citep{crest2025} is omitted because it identifies reflection-related attention heads rather than a single positive--negative reflection direction; Random is an isotropic hidden-space reference for negligible structured overlap. Median overlap is $0.0332$ for the raw direction and $0.0385$ after PCA, compared with $0.00048$ for Random. Orthogonalization reduces it to $0.00042$. Thus, orthogonalization removes most of the measured reasoning-proxy overlap; PCA alone does not.

\paragraph{Layer calibration and projection removal.}
Table~\ref{tab:ablation}, Panel B tests layer calibration and the choice of projection removal over additive steering. Steering all candidate layers adds only 1.2 points of token reduction but lowers accuracy by 4.0 points, while an additive update increases tokens by 27.8\%. Calibration therefore decides \emph{where} to intervene, and bounded projection removal controls \emph{how}. Together with purification, these choices produce the balanced accuracy--cost behavior of the full method. This answers RQ2.

\subsection{RQ3: Does Reflection Steering transfer across models and benchmarks?}
To answer this question, Table~\ref{tab:transferstress} re-runs the same pipeline with model-specific calibration, while Table~\ref{tab:main} reuses the fixed 30B controller across tasks.

\begin{table}[t]
\centering
\small
\setlength{\tabcolsep}{2.8pt}
\begin{tabular*}{\columnwidth}{@{\extracolsep{\fill}}lllrrr@{}}
\toprule
Model & Benchmark & Method & Tok.\,$\downarrow$ & $\Delta$Tok.\,(\%)\,$\uparrow$ & Acc.\,$\uparrow$\\
\midrule
\multirow{3}{*}{Qwen3-8B} & \multirow{3}{*}{MATH-500} & \rawref{Raw model} & \rawref{4345} & \rawref{---} & \rawref{93.6}\\
 & & ReflCtrl & 4449 & $-2.4$ & \best{93.6}\\
 & & \best{RS} & \best{4060} & \best{6.6} & 93.2\\
\midrule
\multirow{3}{*}{Qwen3-8B} & \multirow{3}{*}{GPQA-D} & \rawref{Raw model} & \rawref{8485} & \rawref{---} & \rawref{57.6}\\
 & & ReflCtrl & 8292 & 2.3 & 56.1\\
 & & \best{RS} & \best{7670} & \best{9.6} & \best{56.1}\\
\midrule
\multirow{3}{*}{QwQ-32B} & \multirow{3}{*}{MATH-500} & \rawref{Raw model} & \rawref{3617} & \rawref{---} & \rawref{75.6}\\
 & & ReflCtrl & 2945 & 18.6 & 74.2\\
 & & \best{RS} & \best{2662} & \best{26.4} & \best{74.8}\\
\midrule
\multirow{3}{*}{QwQ-32B} & \multirow{3}{*}{GPQA-D} & \rawref{Raw model} & \rawref{7599} & \rawref{---} & \rawref{65.2}\\
 & & ReflCtrl & 7066 & 7.0 & 63.1\\
 & & \best{RS} & \best{6401} & \best{15.8} & \best{63.1}\\
\bottomrule
\end{tabular*}
\caption{\textbf{Cross-model results.} RS denotes Reflection Steering. All rows use matched thinking-token accounting.}
\label{tab:transferstress}
\end{table}

\paragraph{Cross-model applicability.}
Across Qwen3-8B and QwQ-32B, Reflection Steering continues to reduce thinking tokens. It saves 6.6\% on MATH-500 and 9.6\% on GPQA-D with 8B, and 26.4\% and 15.8\% with 32B, respectively. At the evaluated settings, these reductions exceed those of the reproduced ReflCtrl comparator in all four settings. The number of selected layers, the best $\alpha$, and the achievable compression remain model-dependent.

\paragraph{Cross-Benchmark transfer.}
The 30B direction and calibration are learned from mathematical reasoning, then reused without refitting on GPQA-Diamond. The same controller still reduces tokens on this scientific-reasoning benchmark, supporting a reflection-associated representation that transfers beyond the construction task.

\begin{figure}[t]
\centering
\includegraphics[width=0.98\columnwidth]{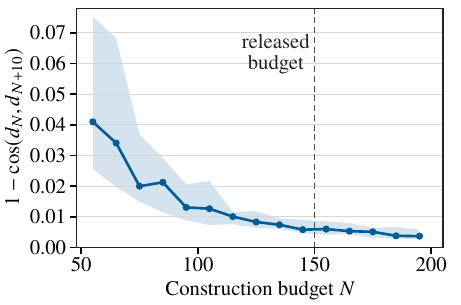}
\caption{\textbf{Direction stability versus construction budget.} Median $1-\cos(d_N,d_{N+10})$ over ten random orderings; the band spans the 5th--95th percentile.}
\label{fig:construction}
\end{figure}

\paragraph{Direction-construction stability.}
Figure~\ref{fig:construction} shows that the direction changes less as the construction budget grows. The $140{\rightarrow}150$ and $150{\rightarrow}160$ cosines are $0.9942$ and $0.9940$, and $N=150$ matches $N=200$ at median cosine $0.9761$. We therefore use $N=150$ as an empirical operating point in this flatter region, rather than as a proven sufficient budget.

\paragraph{Calibration-selected layers.}
Leave-one-out calibration yields an eight-layer stability core (L36, L38, L39, L41--L45). On GPQA-Diamond seeds 0 and 128 at $\alpha=0.7$, it retains 16.1\% token reduction with zero collapse and a $-1.52$-point mean accuracy change. Thus, the effect is not tied to one calibration split. Overall, RQ3 shows that the framework transfers across tasks and checkpoints, while the selected layers and operating point remain model-specific.

\section{Conclusion}
In this paper, we introduce Reflection Steering, a training-free residual-stream intervention combining direction purification, layer calibration, and bounded projection removal. Across three Qwen-family checkpoints and two benchmarks, it reduces thinking tokens in all six matched settings by 16.9\% on average. Accuracy preservation is strongest on mathematical reasoning, while GPQA-Diamond shows cross-task token reduction without established one-point equivalence. Overall, the results support tunable activation-space control without weight updates.

Future work will extend Reflection Steering to broader model families, reasoning tasks, and decoding settings. It will also explore adaptive controllers that vary intervention strength across layers and decoding steps, aiming to remove redundant rechecking while preserving useful verification and correction.

\begingroup
\small
\bibliography{references}
\endgroup

\end{document}